\documentclass{article}

\usepackage[numbers]{natbib}

\usepackage[acronym,smallcaps,nowarn]{glossaries}
\glsdisablehyper{}

\newacronym{RL}{rl}{reinforcement learning}
\newacronym{SIL}{sil}{self-imitation learning}
\newacronym{ES}{es}{evolutionary strategies}
\newacronym{PBT}{pbt}{population-based training}
\newacronym{CEM}{cem}{cross-entropy method}
\newacronym{PPO}{ppo}{proximal policy optimization}
\newacronym{TD3}{td3}{twin-delayed deep deterministic policy gradient}
\newacronym{DDPG}{ddpg}{deep deterministic policy gradient}

\newacronym{OHT}{oht}{online hyper-parameter tuning}

\newacronym{EM}{em}{expectation maximization}
\newacronym{IS}{is}{importance sampling}

\newacronym{HER}{her}{hindsight experience replay}
\newacronym{HPG}{hpg}{hindsight policy gradient}
\newacronym{hEM}{$\mathrm{h}$em}{hindsight expectation maximization}

\newacronym{MDP}{mdp}{markov decision process}
\newacronym{Variational RL}{variational rl}{variational reinforcement learning}

\newacronym{ELBO}{elbo}{evidence lower bound}

\usepackage[nonatbib,preprint]{nips_2020}
\usepackage[utf8]{inputenc} 
\usepackage[T1]{fontenc}    
\usepackage{hyperref}       
\usepackage{url}            
\usepackage{booktabs}       
\usepackage{amsfonts}       
\usepackage{nicefrac}       
\usepackage{microtype}      
\usepackage{amsmath}
\usepackage{algorithm,algorithmic}
\usepackage{graphicx}
\usepackage{amsmath,amssymb,amsthm}
\usepackage{algorithm,algorithmic}
\usepackage{subfigure}

\usepackage{xcolor}
\usepackage{framed}
\usepackage{graphicx}
\usepackage{mathtools}
\graphicspath{ {images/} }
\usepackage{tikz}
\usetikzlibrary{arrows,shapes,snakes,automata,backgrounds,petri}
\usepackage{pdfpages}
\usepackage{hyperref}
\hypersetup{
	colorlinks=true,
	linkcolor=blue,
	filecolor=magenta,      
	urlcolor=cyan,
}

\usepackage[T1]{fontenc}
\usepackage[utf8]{inputenc}
\usepackage{tabularx,ragged2e,booktabs,caption}
\newcolumntype{C}[1]{>{\Centering}m{#1}}

\usepackage{caption}

\usepackage{enumitem}

\usepackage{thmtools}
\usepackage{thm-restate}
\usepackage{hyperref}
\usepackage{cleveref}

\newtheorem{predefinition}{Definition}

\newtheorem{theorem}{Theorem}

\newtheorem{preproposition}{Proposition}
\newenvironment{proposition}[1]
{
\begin{preproposition}
}{
\end{preproposition}
}

\usepackage{mathtools}

\author{%
  Yunhao Tang \\
  Columbia University\\
  \texttt{yt2541@columbia.edu} \\
  \And
  Krzysztof Choromanski \\
  Google Brain Robotics \\
  \texttt{kchoro@google.com} \\
}

\title{Online Hyper-parameter Tuning in Off-policy Learning via Evolutionary Strategies}

\begin{document}

\maketitle

\begin{abstract}
Off-policy learning algorithms have been known to be sensitive to the choice of hyper-parameters. However, unlike near on-policy algorithms for which hyper-parameters could be optimized via e.g. meta-gradients, similar techniques could not be straightforwardly applied to off-policy learning. In this work, we propose a framework which entails the application of Evolutionary Strategies to online hyper-parameter tuning in off-policy learning. Our formulation draws close connections to meta-gradients and leverages the strengths of black-box optimization with relatively low-dimensional search spaces. We show that our method outperforms state-of-the-art off-policy learning baselines with static hyper-parameters and recent prior work over a wide range of continuous control benchmarks.
\end{abstract}

\section{Introduction}

Off-policy learning is a powerful paradigm for \gls{RL} problems. Despite its great promise, when combined with neural networks in many modern applications \citep{mnih2013,silver2016}, off-policy learning suffers from constant instability, also partly characterized as the deadly triad \citep{sutton2018reinforcement,van2018deep}. As a result, additional empirical techniques must be implemented to achieve more robust performance in practice, e.g. target networks \citep{mnih2013}. Though theory suggests that off-policy learning could be performed between highly different behavior policy $\mu$ and target policy $\pi$, in challenging domains the best performance is obtained when data are near on-policy, i.e. $\pi \approx \mu$ \citep{kapturowski2018recurrent}. In batch \gls{RL}, an extreme special case of off-policy learning, where the data are collected under a behavior policy $\mu$ before hand and no further data collection is allowed, naive applications of off-policy algorithms do not work properly \citep{fujimoto2018off}. In addition to algorithmic limitations, the search of good hyper-parameters for off-policy algorithms is also critical yet brittle. For example, prior work has observed that the performance is highly sensitive to hyper-parameters such as learning rates and depends critically on seemingly heuristic techniques such as $n$-step updates \citep{lillicrap2015continuous,barth2018distributed,hessel2018rainbow}.

In this work, we focus on this latter source of instability for off-policy learning, i.e. hyper-parameter tuning. Unlike supervised learning counterparts, where a static set of hyper-parameters might suffice, for general \gls{RL} problems it is desirable to adapt the hyper-parameters \emph{on the fly} as the training procedures are much more non-stationary. Though it is possible to design theoretically justified scheme for hyper-parameters, such methods are usually limited to a set of special quantities, such as the eligibility trace $\lambda$ \citep{mann2016adaptive} or mixing coefficient $\alpha$ for alpha-retrace \citep{rowland2019adaptive}. More generally, the tuning of generic hyper-parameters could be viewed as greedily optimizing certain meta-objectives at each iteration \citep{xu2018meta,paul2019fast,zahavy2020self}. For example, in near on-policy algorithms such as IMPALA \citep{espeholt2018impala}, hyper-parameters are updated by meta-gradients \citep{xu2018meta,zahavy2020self} (in such literature, trainable hyper-parameters are called \emph{meta-parameters}), which are calculated via back-propagation from the meta-objectives.

However, in off-policy learning, techniques such as meta-gradients are not immediately feasible. Indeed, since the existing formulation of meta-gradients \citep{xu2018meta,zahavy2020self} is limited to near on-policy actor-critic algorithms \citep{mnih2016,espeholt2018impala}, its extension to replay-based off-policy algorithms is not yet clear. The difficulty arises from the design of many off-policy algorithms - many off-policy updates are not based on the target \gls{RL} objective but proxies such as Bellman errors \citep{mnih2013,lillicrap2015continuous} or off-policy objectives \citep{degris2012}. This makes it challenging to define and calculate meta-gradients, which requires differentiating through the \gls{RL} objectives via policy gradients \citep{xu2018meta}. To adapt hyper-parameters in such cases, a naive yet straightforward resort is to train multiple agents with an array of hyper-parameters in parallel as in \gls{PBT}, and update hyper-parameters with e.g. genetic algorithms \citep{jaderberg2017population}. Though being more blackbox in nature, \gls{PBT} proved high-performing yet too costly in practice.

\begin{figure}[h]
\centering
\subfigure[\textbf{Mean}]{\includegraphics[width=.32\linewidth]{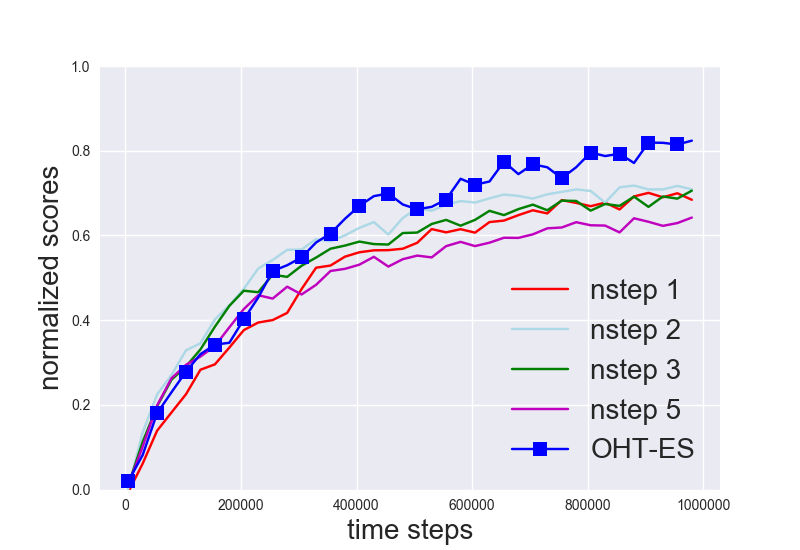}}
\subfigure[\textbf{Median}]{\includegraphics[width=.32\linewidth]{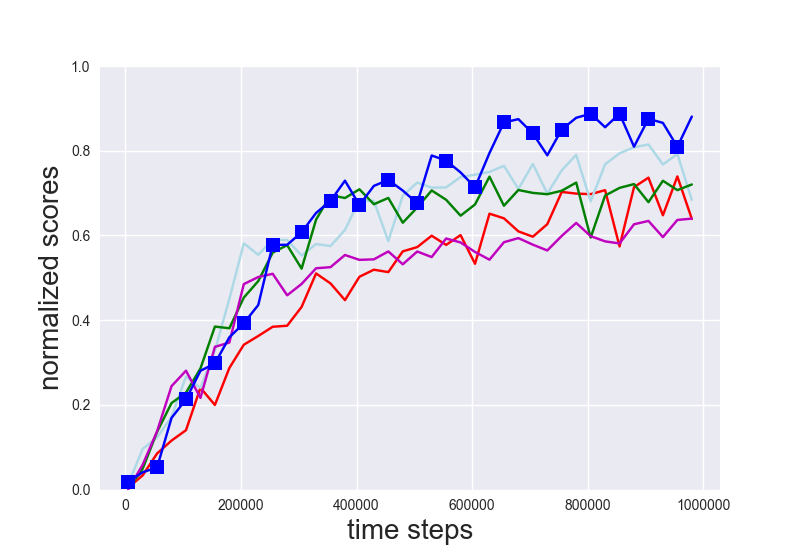}}
\subfigure[\textbf{Best ratio}]{\includegraphics[width=.32\linewidth]{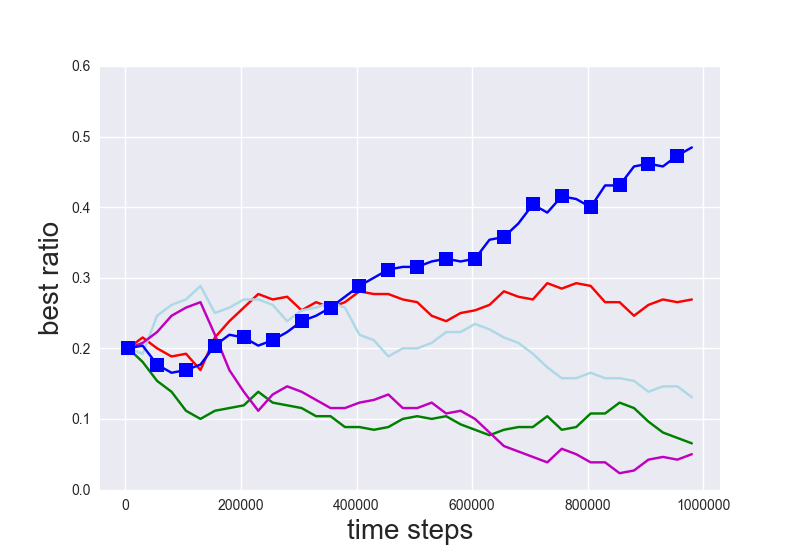}}
\caption{\small{Training performance of discrete hyper-parameter adaptation on control suite tasks. Each plot shows a separate performance statistics during training (mean, median and win ratio). The statistics are normalized per task and averaged over $13$ simulated locomotion tasks. Observe that ES adaptation outperforms other baselines in every performance metric. See Section \ref{sec:experiment}  and Appendix \ref{appendix:exp} for detailed descriptions of the normalized scores.}}
\label{figure:discrete-statistics}
\end{figure}

\paragraph{Main idea.} We propose a framework for optimizing hyper-parameters within the lifetime of a single agent (unlike multiple copies in \gls{PBT}) with \gls{ES}, called \gls{OHT}-\gls{ES}. \gls{ES} are agnostic to the off-policy updates of the baseline algorithm and can readily adapt discrete/continuous hyper-parameters effectively. With the recent revival of \gls{ES} especially for low-dimensional search space \citep{salimans2017evolution,ha2018world}, we will see that our proposal combines the best of both off-policy learning and \gls{ES}. 

\gls{OHT}-\gls{ES} outperforms off-policy baselines with static hyper-parameters. In Figure \ref{figure:discrete-statistics}, we show the significant performance gains of off-policy learning baselines combined with \gls{OHT}-\gls{ES} (blue curves), compared to static hyper-parameters. We evaluate all algorithms with normalized scores over 13 simulated control tasks (see Section \ref{sec:experiment} for details). The performance gains of \gls{OHT}-\gls{ES} are consistent across all three reported metrics over normalized scores.

\section{Background}

In the standard formulation of \gls{MDP}, at a discrete time $t\geq 0$, an agent is in state $x_t\in \mathcal{X}$, takes action $a_t\in\mathcal{A}$, receives a reward $r_t\in \mathbb{R}$ and transitions to a next state $x_{t+1}\sim p(\cdot|x_t,a_t)\in \mathcal{S}$. A policy $\pi(a|x):\mathcal{S}\mapsto \mathcal{P}(\mathcal{A})$ defines a map from states to distributions over actions. The standard objective of \gls{RL} is to maximize the expected cumulative discounted returns $J(\pi) \coloneqq \mathbb{E}_\pi[\sum_{t\geq 0}\gamma^t r_t]$ with a discount factor $\gamma\in (0,1)$.

\subsection{Off-policy learning}

Off-policy learning entails policy optimization through learning from data generated via arbitrary behavior policy, e.g. historical policies. For example, Q-learning \citep{watkins1992q} is a prominent framework for off-policy learning, where given a $(n+1)$-step partial trajectory $(x_i,a_i,r_i)_{i=0}^{n}$, the $n$-step Q-learning optimizes a parameterized Q-function $Q_\theta(x,a)$ by minimizing the Bellman error 
\begin{align*}
\min_\theta\ \mathbb{E}_{\mathcal{D}}[(Q_\theta(x_0,a_0) - Q_{\text{target}})^2],
\end{align*}
where $Q_{\text{target}} \coloneqq \sum_{i=0}^{n-1}\gamma^i r_i + \gamma^n \max_{a^\prime} Q_\theta(x_n,a^\prime)$ is the $n$-step target and  $\mathbb{E}_\mathcal{D}[\cdot]$ denotes that the data are sampled from a replay buffer $\mathcal{D}$. When $n=1$, Q-learning converges to the optimal solution in tabular cases and under mild conditions \citep{watkins1992q}. Recently, \citep{rowland2019adaptive} shows that general uncorrected $n$-step updates for $n\geq 2$ introduce target bias in exchange for fast contractions to the fixed point, which tends to bring empirical gains. Though there is no general optimality guarantee for $n\geq2$, prior work finds that employing $n\geq 2$  significantly speeds up the optimization in challenging image-based benchmark domains \citep{mnih2016,hessel2018rainbow,kapturowski2018recurrent}. 
Other related prominent off-policy algorithms include off-policy policy gradients \citep{degris2012,wang2016}, whose details we omit here.

\subsection{Off-policy actor-critic}
By construction, Q-learning requires a maximization over actions to compute target values, which becomes intractable when the action space is continuous, e.g. $\mathcal{A}=[-1,1]^m$. To bypass such issues, consider a policy  $\pi_\phi(\cdot|x)$ as an approximate maximizer, i.e. $\pi_\phi(x) \approx \arg\max_a Q_\theta(x,a)$. This produces the Q-function target $Q_{\text{target}} = \sum_{i=0}^{n-1}\gamma^i r_i + \gamma^n  Q_\theta(x_n,\pi_\phi(x_n))$. The Q-function (critic) and the policy (actor) are alternately updated as follows, with learning rate $\alpha$,
\begin{align}
    \theta^\prime &\leftarrow \theta - \alpha \nabla_\theta \mathbb{E}_{\mathcal{D}}[(Q_\theta(x_0,a_0) - Q_{\text{target}})^2],
    \ \ \phi^\prime \leftarrow \phi + \alpha \nabla_\phi \mathbb{E}_{\mathcal{D}}[Q_\theta(x,\pi_\phi(x))]. \label{eq:ddpg}
\end{align}
Depending on whether the actor $\pi_\phi(x)$ or the critic $Q_\theta(x,a)$ is fully optimized at each iteration, there are two alternative interpretations of the updates defined in Eqn.(\ref{eq:ddpg}). When the policy is fully optimized such that $\pi_\phi(x) = \arg\max Q_\theta(x,a)$, the updates are exact $n$-step Q-learning. When the critic is fully optimized such that $Q_\theta(x,a) = Q^{\pi_\phi}(x,a)$, the updates are $n$-step SARSE for policy evaluation with deterministic policy gradients \citep{silver2014}. In practice, critic and actor updates take place alternately and the algorithm is a mixture between value iteration and policy iteration \citep{sutton1998}. Built upon the updates (Eq.(\ref{eq:ddpg})), additional techniques such as double critic \citep{fujimoto2018addressing} and maximum entropy formulation \citep{haarnoja2018soft} could greatly improve the stability of the baseline algorithm.

\subsubsection{Evolutionary strategies}
\gls{ES} are a family of zero-order optimization algorithms (see e.g. \citep{hansen2003reducing,de2005tutorial,wierstra2014natural,salimans2017evolution}), which have seen recent revival for applications in \gls{RL} \citep{salimans2017evolution}. In its generic form, consider a function $f(\theta)$ with parameter $\theta$, the aim is to optimize $\max_\theta f(\theta)$ with only queries of the function values. For simplicity, assume $\theta$ is continuous and consider the \gls{ES} gradient descent formulation introduced in \citep{salimans2017evolution}. Instead of optimizing $f(\theta)$ directly, consider a smoothed objective $F(\theta) = \mathbb{E}_{x\sim\mathcal{N}(\theta,\sigma^2\mathbb{I})} [f(x)]$ with some fixed variance parameter $\sigma^2$. It is then feasible to approximate the gradient $\nabla_\theta F(\theta)$ with $N$-sample unbiased estimates, in particular,
$\hat{g}_\theta = \frac{1}{N} \sum_{i=1}^N f(\theta + \sigma \epsilon_i) \frac{\epsilon_i}{\sigma} \approx \nabla_\theta F(\theta)$ where $\epsilon_i \sim \mathcal{N}(0,\mathbb{I})$ are i.i.d. Gaussian vectors. A naive approach to \gls{RL} is to flatten the sequential problem into a one-step blackbox problem, by setting $f(\theta) \coloneqq J(\pi_\theta)$. Despite its simplicity, this approach proved efficient compared to policy gradient algorithms \citep{salimans2017evolution,choromanski2018structured,mania2018simple}, though generally its sample efficiency could not match that off-policy algorithms.

\section{Online Hyper-parameter Tuning via Evolutionary Strategies}


Let $\eta$ denote the set of adjustable real-valued hyper-parameters, e.g. the learning rate $\alpha$ and a probability distribution over $n$-step targets for discrete $n$. At iteration $t$ with actor-critic parameter $\psi_t = (\theta_t,\phi_t)$, given replay buffer $\mathcal{D}$, the algorithm constructs an update $f(\psi_t, \mathcal{D}, \eta_t)$ such that $\psi_{t+1} = \psi_t + f(\psi_t,\mathcal{D},\eta_t)$ \citep{xu2018meta}. Here we make explicit the dependency of the update function $f$ on the replay $\mathcal{D}$. For example, the update function $f$ could be the gradient descents defined in Eqn.(\ref{eq:ddpg}).

When the algorithm does not update hyper-parameter at all $\eta_t \equiv \eta$, and we reduce to the case of static hyper-parameters. One straightforward way to update the hyper-parameter is to greedily optimize the hyper-parameters against some meta objective $L(\psi,\eta)$ \citep{xu2018meta,paul2019fast}, such that
\begin{align}
    \eta_{t+1} = \arg\max_\eta L(\psi_{t+1},\eta),\ \text{s.t.}\ \   \psi_{t+1} = \psi_t + f(\psi_t,\mathcal{D},\eta). \label{eq:greedymeta}
\end{align}
Since the motivation of hyper-parameter adaptation was to better optimize the \gls{RL} objective, it is natural to set the meta objective as the target \gls{RL} objective, i.e. cumulative returns $L(\psi,\eta) \coloneqq J(\pi_\phi)$. 

\subsection{Methods}

 \begin{algorithm}[t]
	\begin{algorithmic}[1]
		\STATE Input: off-policy update function $f(\psi,\mathcal{D},\eta)$ and agent parameter $\psi$. 
		\WHILE { $t=0,1,2...$ }
		\STATE Sample $N$ hyper-parameters from a Gaussian distribution $\eta_t^{(j)} \sim \mathcal{N}(\mu_t,\sigma^2),1\leq j\leq N$.
		\STATE Train $N$ off-policy agents: $\psi_{t+1}^{(j)} \leftarrow \psi_t^{(j)} + f(\psi_t^{(j)},\mathcal{D},\eta_t^{(j)}),1\leq j\leq N$.
		\STATE Collect rollout with agent parameter $\psi_t^{(j)}$, save data to $\mathcal{D}$. Estimate $\hat{L}_t^{(j)},1\leq j\leq N$.
		\STATE Update the hyper-parameter distribution based on Eqn.(\ref{eq:esmeta}).
		\ENDWHILE
	\end{algorithmic}
	 	\caption{Online Hyper-parameter Tuning via Evolutionary Strategies (\gls{OHT}-\gls{ES})}\label{algo:routine}
\end{algorithm}

Now we describe Online Hyper-parameter Tuning via Evolutionary Strategies (\gls{OHT}-\gls{ES}). Note that the framework is generic as it could be combined with any off-policy algorithms with update function $f$. Recall that the update function returns a new parameter $\psi^\prime = \psi + f(\psi,\mathcal{D},\eta)$. The general meta algorithm is presented in Algorithm \ref{algo:routine}, where we assume hyper-parameters to be real-valued. It is straightforward to derive similar algorithms for discrete hyper-parameters as explained below.

Consider at iteration $t$ of learning, the agent maintains a parametric distribution over hyper-parameters, e.g. Gaussian $\mathcal{N}(\mu,\sigma^2)$ with tunable mean $\mu$ and fixed variance $\sigma^2$. Then we sample a population of $N$ actor-critic agents $\{\psi_t^{(j)}\}_{j=1}^N$ each with a separate hyper-parameter $\{\eta_t^{(j)}\}_{j=1}^N$ drawn from the parametric distribution $\eta^{(j)} \sim \mathcal{N}(\mu, \sigma^2)$. Then for each of the $N$ copies of the agent, we update their parameters via the off-policy subroutine $\psi_{t+1}^{(j)} = \psi_t^{(j)} + f(\psi_t^{(j)},\eta_t^{(j)}),\forall j$. After the update is complete, each agent with parameter $\psi_{t+1}^{(j)}$ collects rollouts from the environment and saves the data to $\mathcal{D}$. From the rollouts, construct estimates of the meta objective $\hat{L}_t^{(j)}\coloneqq\hat{L}(\pi_t^{\phi^{(j)}},\eta_t^{(j)})$. Fianlly, the hyper-parameter mean $\mu$ is updated via a \gls{ES} subroutine. For example, we might apply \gls{ES} gradient ascent \citep{salimans2017evolution} and the new distribution parameter is updated with learning rate $\beta$,
\begin{align}
    \mu_{t+1}\leftarrow \mu_t + \beta \frac{1}{\sigma N} \sum_{j=1}^{N}\hat{L}_t^{(j)} (\eta_t^{(j)} - \mu_t) 
    \label{eq:esmeta}
\end{align} 

\paragraph{Discrete hyper-parameters.} We also account for the case where the hyper-parameters take values from a discrete set of $K$ values, denoted as $\eta \in \{1,2...K\}$. In such cases, instead of maintaining a parametric Gaussian distribution over hyper-parameters such that $\eta^{(j)} \sim \mathcal{N}(\mu,\sigma^2)$, we maintain a categorical distribution $\eta \sim \text{Cat}(L)$ where $L \in \mathbb{R}^K$ is the logits and $P(\eta=i) \equiv \text{softmax}(L_i)$. By sampling several hyper-parameter candidates $\eta^{(j)} \sim \text{Cat}(L)$, we could construct a score function gradient estimator \citep{williams1992} for the logits $L$
\begin{align}
    L_{t+1} \leftarrow L_t + \beta \frac{1}{N} \sum_{j=1}^N \hat{L}_t^{(j)} \nabla_L \log P(\eta=\eta^{(j)}).
    \label{eq:discrete-esmeta}
\end{align}

\subsection{Connections to prior work}

We make explicit the connections between our approach and closely related prior work. 

\paragraph{Connections to meta-gradients.} When hyper-parameters are real-valued, the \gls{ES} updates defined in Eqn.(\ref{eq:esmeta}) closely relates to meta-gradients \citep{xu2018meta}, as summarized in the following proposition.
\begin{proposition}

(Proved in Appendix \ref{appendix:theory}) Assume that sampled hyper-parameters $\eta$ follow a Gaussian distribution $\eta\sim\mathcal{N}(\mu,\sigma^2)$. Then the following holds,
\begin{align}
    \lim_{\sigma\rightarrow 0}\mathbb{E}[\frac{1}{\sigma N} \sum_{j=1}^{N}\hat{L}_t^{(j)} (\eta_t^{(j)} - \mu_t)] = [\nabla_\psi L(\psi,\mu)]_{\psi=\psi_t+f(\psi_t,\mathcal{D},\mu)} [\nabla_\mu f(\psi_t,\mathcal{D},\mu)]_{\mu=\mu_t}. 
    \label{eq:equiv}
\end{align}
\end{proposition} 
 Since \gls{ES} gradient updates are a zero-order approximation to the analytic gradients, this connection should be intuitive. Note that the RHS of Eq.(\ref{eq:equiv}) differs from meta-gradient updates in practice in several aspects \citep{xu2018meta}: in general, meta-gradients could introduce trace parameters to stabilize the update, and the gradient $\nabla_\psi L(\psi)$ is evaluated at $\psi=\psi_t$ instead of $\psi_t+f(\psi_t,\mathcal{D},\mu)$ as defined above.

\paragraph{Connections to near on-policy methods.} For near on-policy algorithms such as A2C, TRPO and PPO \citep{mnih2016,schulman2015,schulman2017}, there are natural constraints on the parameter updates. As a result, given the meta objective $L(\psi,\eta)=J(\pi_\phi)$ of one hyper-parameter value $\eta$, it is possible to estimate meta objectives $L(\psi^\prime,\eta^\prime)$ at alternative hyper-parameter values $\eta^\prime$  with importance sampling (IS) \citep{paul2019fast}. Then meta objectives could be greedily optimized via even zero-order methods. However, it is not clear how correlations/variance of such IS-\gls{ES}timated meta objectives impact the updates, as they are estimated from the same data. Alternative to IS, we estimate $L(\psi,\eta)$ via the Monte-Carlo sample of cumulative returns under $(\psi,\eta)$, which is applicable when trust regions are not available (as with many off-policy algorithms) and when policies are deterministic \citep{lillicrap2015continuous,fujimoto2018addressing}.

\paragraph{Connections to \gls{ES}-\gls{RL}.} Our method closely relates to prior work on combining \gls{ES} with gradient based off-policy \gls{RL} algorithms \cite{khadka2018evolution,pourchot2018cem}, which we name \gls{ES}-\gls{RL}. These algorithms maintain a population of off-policy agents with parameters $\{\psi^{(j)}\}_{j=1}^N$ and carry out \gls{ES} updates directly on the agent parameter, e.g. genetic algorithm \citep{khadka2018evolution} or cross-entropy method \citep{pourchot2018cem}. They could be interpreted as a special case of our framework: indeed, one could include the trainable agent parameters as part of the hyper-parameter $\eta$ and this formulation reduces to \gls{ES}-\gls{RL}. However, \gls{ES}-\gls{RL} applies \gls{ES} updates to a high-dimensional trainable parameter, which might be less effective than to a low-dimensional hyper-parameter search space. We will examine their relative strengths in Section 4.

\paragraph{Connections to \gls{PBT}.} Our framework could be interpreted as a special variant of \gls{PBT} \citep{jaderberg2017population}, where $N$ copies of the \gls{RL} agents share replay buffers. In particular, \gls{PBT} agents are trained independently in parallel and only exchange information during periodic hyper-parameter updates, while our approach ensures that these $N$ agents share information during training as well. This makes our approach potentially much more sample efficient than \gls{PBT}. It is also worth noting that sharing buffers involves a trade-off - though $N$ agents could utilize others' data for potentially better exploration, the behavior data also become less on-policy for any particular agent and might introduce additional instability \citep{kapturowski2018recurrent}. 

\section{Experiments}
\label{sec:experiment}
In the experiments, we seek to address the following questions: \textbf{(1)} Is \gls{OHT}-\gls{ES} effective for discrete hyper-parameters? \textbf{(2)} Is \gls{OHT}-\gls{ES} effective for continuous hyper-parameters and how does it compare to meta-gradients \citep{xu2018meta}? \textbf{(3)} How is  \gls{OHT}-\gls{ES} compared to highly related methods such as \gls{ES}-\gls{RL} \citep{pourchot2018cem}?

To address \textbf{(1)}, we study the effect of adapting the horizon hyper-parameter $n$ in $n$-step updates. Prior work observed that $n=3$ generally performs well for Atari and image-based continuous control \citep{mnih2016,hessel2018rainbow,barth2018distributed}, though the best hyper-parameter could be task-dependent. We expect  \gls{OHT}-\gls{ES} to be able to adapt to the near optimal hyper-parameters for each task. To address \textbf{(2)}, we study the effect of learning rates, and we compare with an application of meta-gradients \citep{xu2018meta} to off-policy agents. Though prior work focuses on applying meta-gradients to near on-policy methods \citep{xu2018meta}, we provide one extension to off-policy baselines for comparison, with details described below.

\paragraph{Benchmark tasks.} For benchmark tasks, we focus on state-based continuous control. In order to assess the strengths of different algorithmic variants, we consider similar tasks \emph{Walker}, \emph{Cheetah} and \emph{Ant} with different simulation backends from OpenAI gym \citep{brockman2016}, Roboschool \citep{klimov2017roboschool}, DeepMind Control Suite \citep{tassa2018deepmind} and Bullet Physics Engine \citep{coumans2010bullet}. These backends differ in many aspects, e.g. dimensions of observation and action space, transition dynamics and reward functions. With such a wide range of varieties, we seek to validate algorithmic gains with sufficient robustness to varying domains. There are a total of $13$ distinct tasks, with details in Appendix \ref{appendix:exp}.

\paragraph{Base update function.} Since we focus on continuous control, we adopt state-of-the-art TD3 \citep{fujimoto2018addressing} as the baseline algorithm, i.e. the update function $f$ defined in Eqn.(\ref{eq:greedymeta}). 

\subsection{Continuous hyper-parameters}

As an example of adaptive continuous hyper-parameters, we focus on learning rates $\alpha = [\alpha_\pi,\alpha_q]$, which includes the learning rates for actor $\alpha_\pi$ and critic $\alpha_q$ respectively. Extensions to other continuous hyper-parameters are straightforward. For example, the original meta-gradients were designed for discount factor $\gamma$ or eligibility trace $\lambda$ \citep{xu2018meta} and later extended to entropy regularization and learning rates \citep{zahavy2020self}. For baseline TD3, alternative hyper-parameter is the discount $\gamma$, for which we find adaptive tuning does not provide significant gains.

We present results on challenging domains from the DeepMind Control Suite \citep{tassa2018deepmind}, where performance gains are most significant. Detailed environment and  hyper-parameter settings are in Appendix \ref{appendix:exp}. We compare \gls{OHT}-\gls{ES} tuning approach with a variant of meta-gradients: as discussed in Section 3, meta-gradient approaches are less straightforward in general off-policy learning. We derive a meta-gradient algorithm for deterministic actor-critics \citep{lillicrap2015continuous,fujimoto2018addressing} and provide a brief introduction below.

\paragraph{Meta-gradients for deterministic actor-critics.} Deterministic actor-critics maintain a Q-function critic $Q_\theta(x,a)$ and a deterministic actor $\pi_\phi(x)$. We propose to train an alternative critic for policy evaluation $Q_\text{meta}(x,a)\approx Q^{\pi_\phi}$ updated via TD-learning as $Q_\theta(x,a)$. Recall that actor-critics are updated as defined in Eqn.(\ref{eq:ddpg}), and recall $\theta^\prime,\phi^\prime$ to be updated parameters. Next, let the meta objective be the off-policy objective $L(\psi,\eta) \equiv \mathbb{E}[Q^{\pi_\phi}(x,a)] \approx \mathbb{E}[Q_{\text{meta}}(x,a)]$ \citep{degris2012}, where the expectation is taken such that $x\sim\mathcal{D},a=\pi_\phi(x)$. The meta-gradients are calculated as $\Delta \alpha = \mathbb{E}[\nabla_\alpha Q_{\text{meta}}(x,\pi_{\phi^\prime}(x)]$. Please see Appendix \ref{appendix:exp} for a detailed derivation and design choices.

\begin{figure}[h]
\centering
\subfigure[\textbf{DMWalkerRun}]{\includegraphics[width=.24\linewidth]{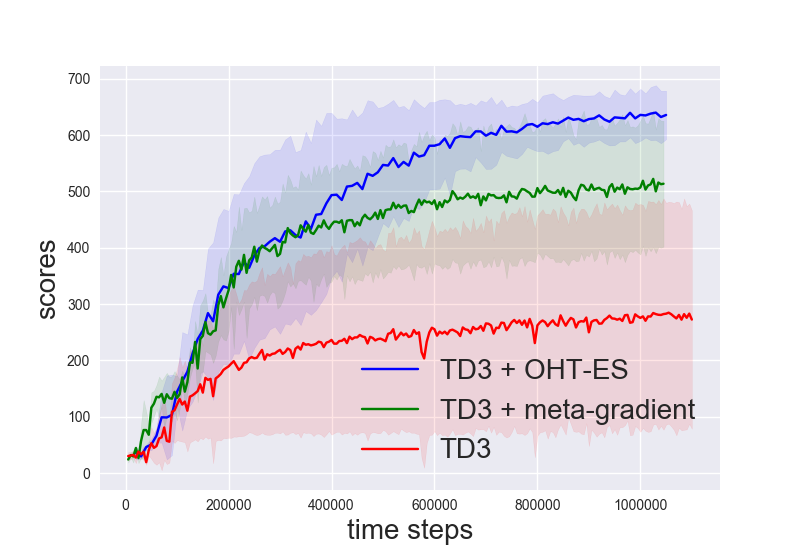}}
\subfigure[\textbf{DMWalkerStand}]{\includegraphics[width=.24\linewidth]{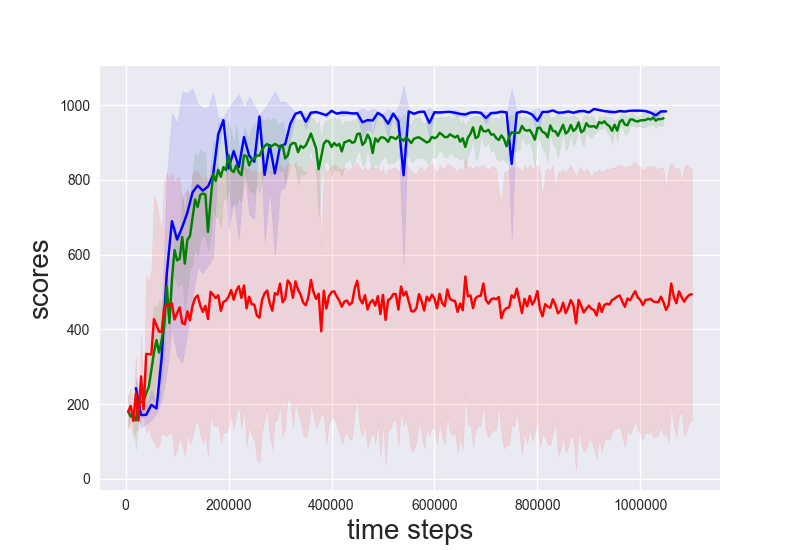}}
\subfigure[\textbf{DMWalkerWalk}]{\includegraphics[width=.24\linewidth]{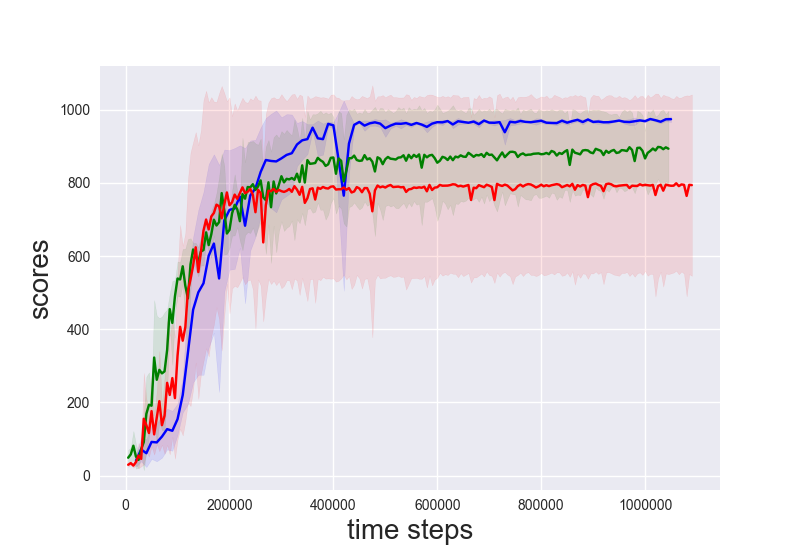}}
\subfigure[\textbf{DMCheetahRun}]{\includegraphics[width=.24\linewidth]{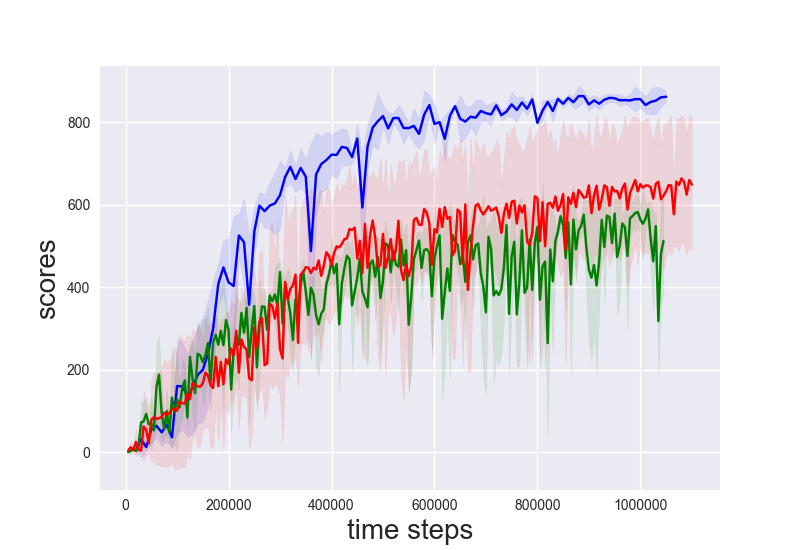}}
\caption{\small{Training performance of continuous hyper-parameter adaptation on control suite tasks. Algorithmic variants are shown in different colors: TD3 (red), meta-gradients $+$ TD3 (green) and \gls{OHT}-\gls{ES} $+$ TD3 (blue). Each task is trained for $10^7$ time steps and each curve shows the $\text{mean} \pm \text{std}$ results across three seeds.}}
\label{figure:cont-dm}
\end{figure}

\paragraph{Evaluations.} The comparisons between \gls{OHT}-\gls{ES}, meta-gradients and TD3 baseline are shown in Figure \ref{figure:cont-dm}. We make a few observations: \textbf{(1)} \gls{OHT}-\gls{ES} consistently achieves the best across all four environments and achieve significant performance gains (asymptotic performance and learning speed) than meta-gradients and TD3; \textbf{(2)} Meta-gradients achieve gains over the baseline most of the time, which implies that there are potentials for improvements due to adaptive learning rate; \textbf{(3)} Baseline TD3 does not perform very well on the control suite tasks. This is in contrast to its high-performance on typical benchmarks such as OpenAI gym \citep{klimov2017roboschool}. This provides strong incentives to test on a wide range of benchmark testbeds in future research as in our paper. We speculate that TD3's suboptimality is due to the fact that its design choices (including hyper-parameters) are not exhaustively tuned on these \emph{new} benchmarks. With adaptive tuning, we partially resolve the issue and obtain performance almost identical to state-of-the-art algorithms on the control suite (e.g. see MPO \citep{abdolmaleki2018maximum}).

\subsection{Discrete hyper-parameters}

\begin{figure}[h]
\centering
\subfigure[\textbf{DMWalkerRun}]{\includegraphics[width=.24\linewidth]{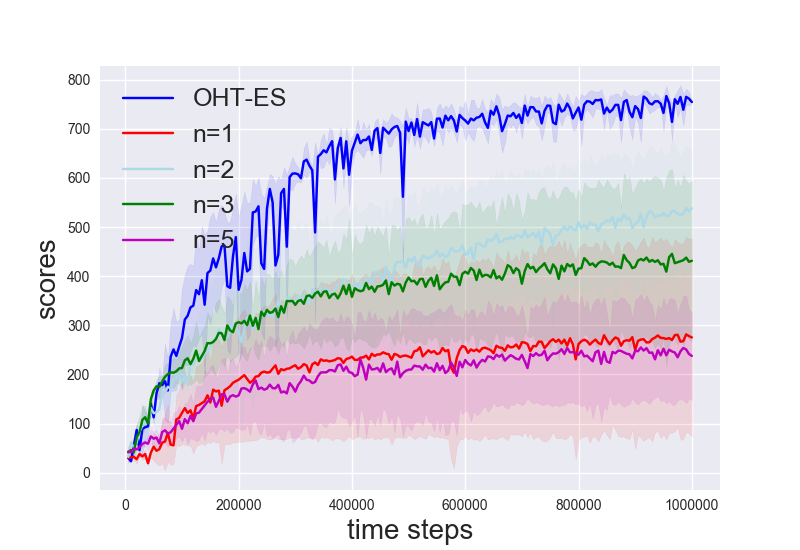}}
\subfigure[\textbf{DMWalkerStand}]{\includegraphics[width=.24\linewidth]{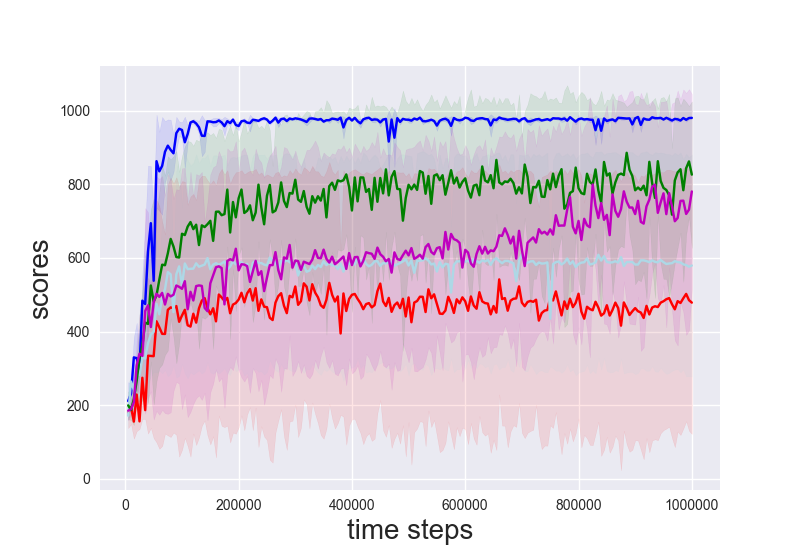}}
\subfigure[\textbf{DMWalkerWalk}]{\includegraphics[width=.24\linewidth]{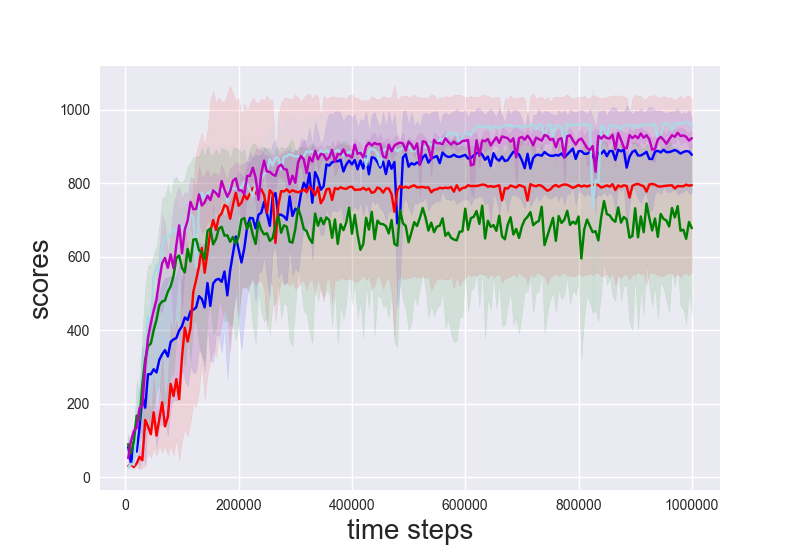}}
\subfigure[\textbf{DMCheetahRun}]{\includegraphics[width=.24\linewidth]{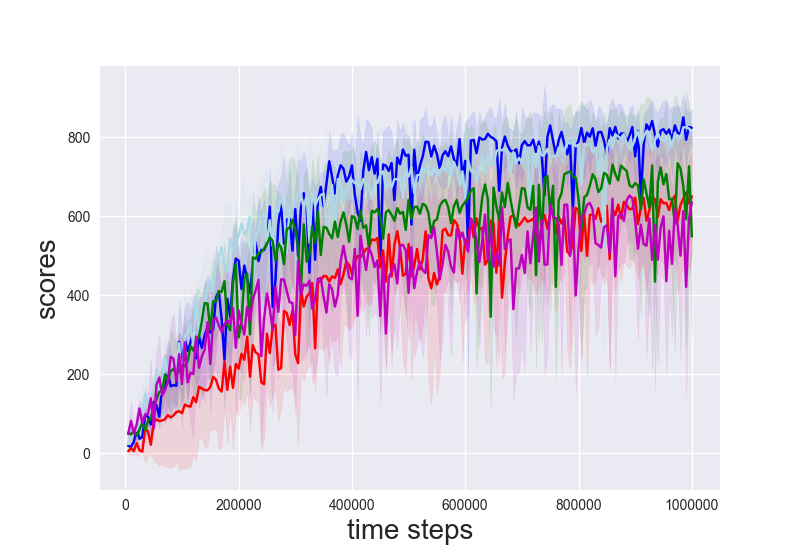}}
\subfigure[\textbf{DMWalkerRun(D)}]{\includegraphics[width=.24\linewidth]{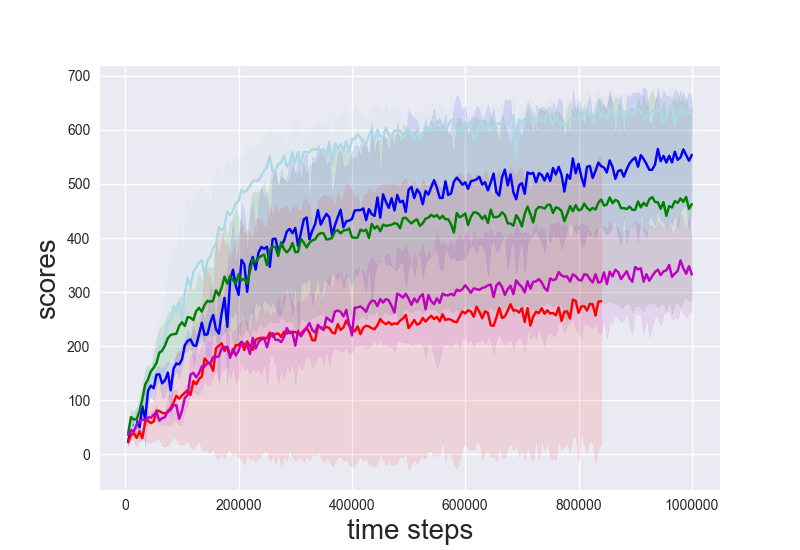}}
\subfigure[\textbf{DMWalkerStand(D)}]{\includegraphics[width=.24\linewidth]{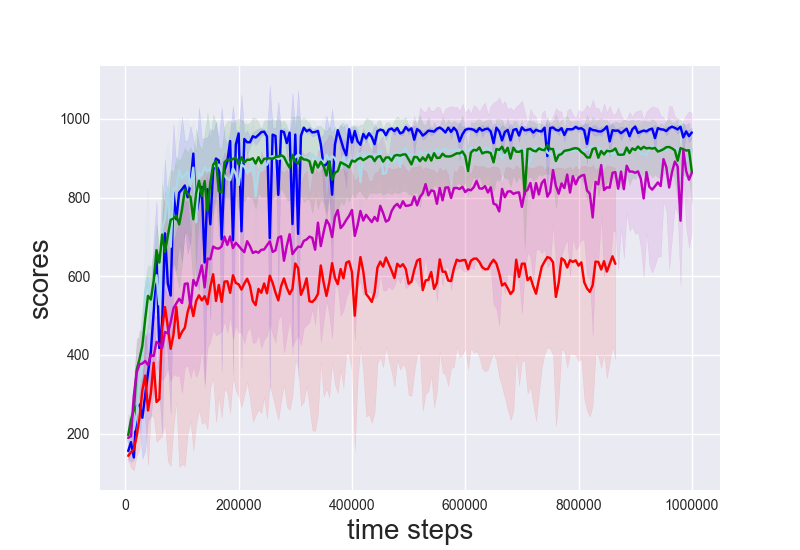}}
\subfigure[\textbf{DMWalkerWalk(D)}]{\includegraphics[width=.24\linewidth]{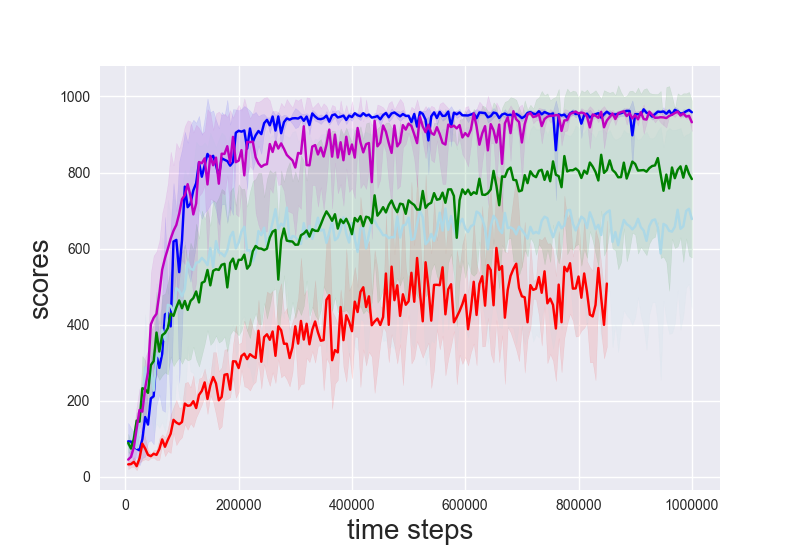}}
\subfigure[\textbf{DMCheetahRun(D)}]{\includegraphics[width=.24\linewidth]{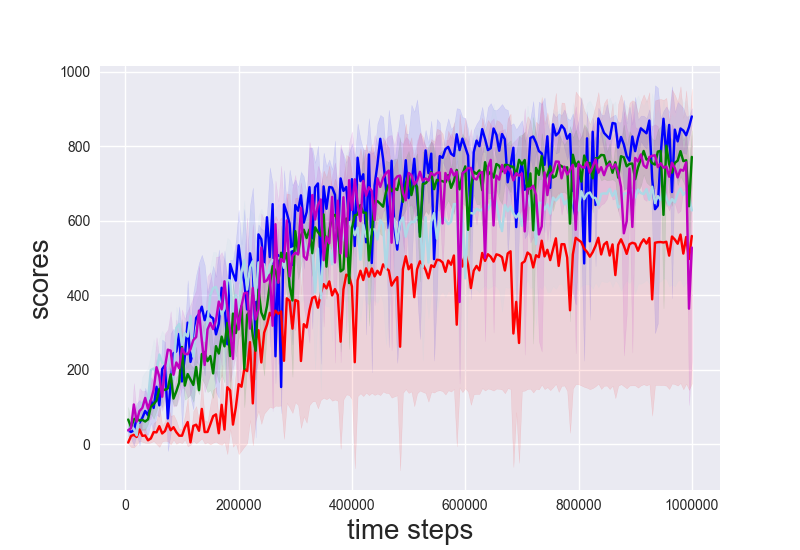}}
\caption{\small{Training performance of discrete hyper-parameter adaptation on control suite tasks. TD3 with different $n$-step parameters are shown in a few colors, while blue shows the result of \gls{ES} adaptation. Each curve shows the $\text{mean} \pm \text{std}$ results across three seeds.}}
\label{figure:discrete-dm}
\end{figure}

As an important example of adaptive discrete hyper-parameters, we focus on the horizon parameter $n$ in $n$-step updates. Due to the discrete nature of such hyper-parameters, it is less straightforward to apply meta-gradients out of the box. As a comparison to the adaptive approach, we consider static hyper-parameters and test if online adaptation brings significant gains. We show results on tasks from the control suite in Figure \ref{figure:discrete-dm} (first row).  For static baselines, we consider TD3 with $n$-step updates with $n=1,2,3,5$.

\paragraph{Evaluation with normalized scores.} Since different tasks involve a wide range of inherent difficulties and reward scales, we propose to calculate the normalized scores for each task and aggregate performance across tasks. This is similar to the standard evaluation technique on Atari games \citep{bellemare2013arcade}. In particular, for each task, let $R_t,1\leq t\leq \tau$ denote the performance curve of a particular algorithm with maximum iteration $\tau$, let $L,U$ be the performance of a random policy and the optimal policy respectively. Then the normalized score is $Z_t = (R_t - L) / (U - L),1\leq t\leq \tau$ and we graph them for comparison (Figure \ref{figure:discrete-statistics}). Please refer to Appendix \ref{appendix:exp} for detailed scores for each tasks.

For convenience, let there be $n$ algorithmic baselines and $m$ tasks. To facilitate comparison of the overall performance, for the $i$-th baseline we calculate the normalized scores for the $j$-th task task $Z_t^{i,j},1\leq t\leq\tau$, and at each time tick $t$ calculate statistics across tasks. There are three statistics: \textbf{mean}, \textbf{median} and \textbf{best ratio}, similar to \citep{rowland2019adaptive}. The best ratio indicates the proportion of tasks on which a certain baseline performs the best. These three statistics summarize the overall algorithmic performance of baseline methods and display their relative strength/weakness.

\paragraph{Evaluations on standard benchmarks.} We present results across all 13 simulated tasks in Figure \ref{figure:discrete-statistics}. In Figure \ref{figure:discrete-dm} (first row), we show detailed training curves on the control suite. Here, \gls{OHT}-\gls{ES} maintains a categorical distribution over $n \in \{1,2,3\}$.

We make several observations from the results: \textbf{(1)} The $n$-step update with $n=1$ achieves better performance across the second largest number of tasks, yet its overall performance is slightly worse than $n=2$ (median). \textbf{(2)} The adaptive $n$-step performs the best across all three metrics. This implies that adaptive $n$-step both achieves significantly better overall performance (mean and median) and achieve the best performance across a considerable proportion of tasks (best ratio); \textbf{(3)} From the best ratio result, we conclude that adaptive $n$-step is able to locate the best $n$-step hyper-parameter for each task through the online adaptation.

\paragraph{Evaluations on delayed reward environment.} Delayed reward environment tests algorithms' capability to tackle delayed feedback in the form of sparse rewards \citep{oh2018self}. In particular, a standard benchmark environment returns dense reward $r_t$ at each step $t$. Consider accumulating the reward over $d$ consecutive steps and return the sum at the end $k$ steps, i.e. $r_t^\prime=0$ if $ t \ \text{mod}\ k \neq 0$ and $r_t^\prime = \sum_{\tau=t-d+1}^t r_\tau$ if $t\ \text{mod}\ d =0$. 

\begin{figure}[h]
\centering
\subfigure[\textbf{Mean}]{\includegraphics[width=.32\linewidth]{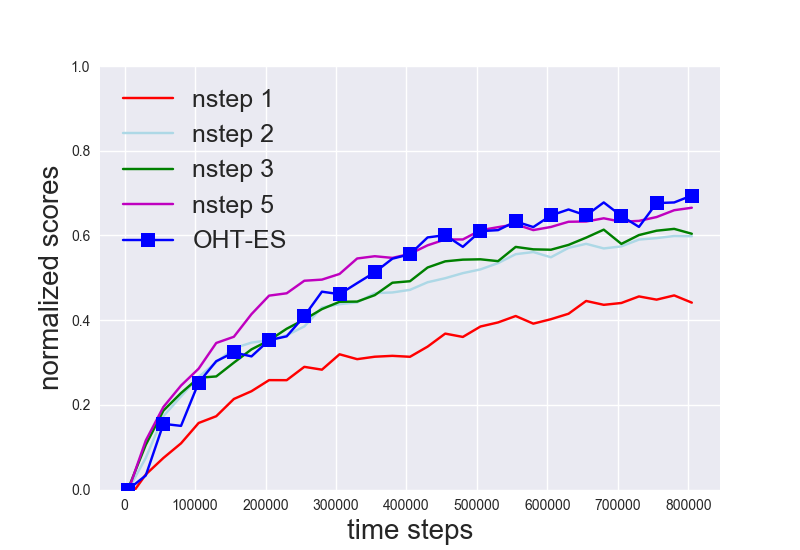}}
\subfigure[\textbf{Median}]{\includegraphics[width=.32\linewidth]{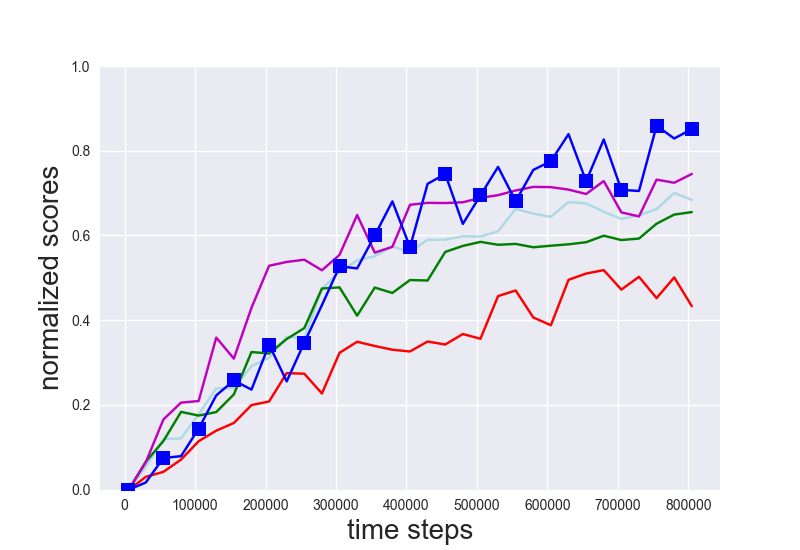}}
\subfigure[\textbf{Best ratio}]{\includegraphics[width=.32\linewidth]{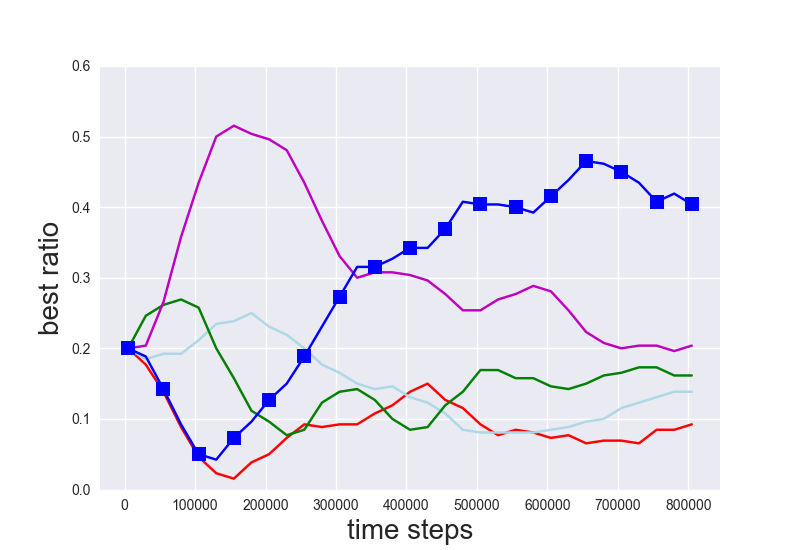}}
\caption{\small{Training performance of discrete hyper-parameter adaptation on control suite tasks with \emph{delayed rewards}. The plot has the exact same setup as Figure \ref{figure:discrete-statistics}.}}
\label{figure:discrete-statistics-delay}
\end{figure}

We present the full results in Figure \ref{figure:discrete-statistics-delay} with normalized scores across all 13 simulated tasks. In Figure \ref{figure:discrete-dm} (bottom row) we show detailed training curves on control suite. Due to delayed rewards, we find it  beneficial to increase the support of the categorical distribution to allow for bootstrapping from longer horizons. As a result, \gls{OHT}-\gls{ES} takes discrete values from $n\in\{1,2,3,4,5\}$. 

We also make several observations: \textbf{(1)} The overall performance of $n$-step update is monotonic in $n$ when $n\leq k$ (mean and median). In particular, we see that when $n=d=5$ the $n$-step update performs the best. Intuitively, we see that $n$-step update skips over $n$ time steps and combine multiple rewards into a single reward, which makes it naturally compatible with the delayed reward signal; \textbf{(2)} The best ratio curves show that $n=d=5$ achieves fastest learning progress across all baselines (including adaptive $n$-step), yet this advantage decays away as the training progresses and adaptive $n$-step takes over. This implies that adapting $n$-step hyper-parameter is critical in achieving more stable long term progress; \textbf{(3)} In terms of overall performance, adaptive $n$-step initially lags behind $n=d=5$ yet quickly catches up and exceeds the latter.

\subsection{Comparison to \gls{ES}-\gls{RL}}

The combination of \gls{ES} with \gls{RL} subroutines have the potential of bringing the best of both worlds. While previous sections have shown that adaptive hyper-parameters achieve generally significantly better performance than static hyper-parameters, how does this approach compare to the case where the \gls{ES} adaptation is applied to the entire parameter vector \citep{khadka2018evolution,pourchot2018cem} ? 

We show results of a wide range of tasks in Table \ref{table:summary}, where we compare several baselines: \gls{ES} adaptation of $n$-step horizon parameter $n$; \gls{ES} adaptation of learning rate $\alpha$; \gls{ES} adaptation to parameter vector (also named \gls{ES}-\gls{RL}) \citep{pourchot2018cem} \footnote{Here, the \gls{ES} update is based on the \gls{CEM} \citep{de2005tutorial} according to \gls{CEM}-\gls{RL} \citep{pourchot2018cem}.}, as well as baseline TD3 and SAC \citep{haarnoja2018soft}. Several observations: \textbf{(1)} Across the selected tasks, \gls{ES} adaptation generally provides performance gains over baseline TD3, as shown by the fact that best performance is usually obtained via \gls{ES} adaptations; \textbf{(2)} \gls{ES} adaptation of hyper-parameters achieve overall better performance than \gls{ES}-\gls{RL}. We speculate that this is partially because \gls{ES}-\gls{RL} naively applies \gls{ES} updates to high-dimensional parameter vectors, which could be highly inefficient. On the other hand, \gls{ES} adaptation of hyper-parameters focus on a compact set of tunable variables and could exploit the strength of \gls{ES} updates to a larger extent.


\begin{table}[t]
\caption{Summary of the performance of algorithmic variants across benchmark tasks. \gls{ES} $+$ $n$-step denotes tuning of $n$-step horizon parameter $n$; \gls{ES} $+$ $\alpha$ denotes tuning of learning rate $\alpha$; \gls{ES} $+$ TD3 denotes the \gls{ES}-\gls{RL} baseline \citep{pourchot2018cem}. For each task, algorithmic variants with top performance are highlighted (multiple are highlighted if they are not statistically significantly different). Each entry shows $\text{mean} \pm \text{std}$ performance.}
\begin{small}
\begin{sc}
\begin{tabular}{C{1.in} *5{C{.7in}}}\toprule[1.5pt]
\bf Tasks & \bf \gls{ES} $+$ $n$-step & \bf \gls{ES} $+$ $\alpha$  & \bf \gls{ES} $+$ TD3 & \bf TD3 & \bf SAC \ \\\midrule
DMWalkerRun & $\mathbf{757 \pm 7}$ & $633 \pm 45$ & $653 \pm 59$ & $274 \pm 200$ & $23 \pm 1$  \\ 
DMWalkerWalk & $881 \pm 107$ & $\mathbf{969 + 3}$ & $\mathbf{966 \pm 5}$ & $793 \pm 241$ & $87 \pm 83$ \\
DMWalkerStand & $\mathbf{981 \pm 3}$ & $\mathbf{985 \pm 3}$ & $\mathbf{984 \pm 3}$ & $487 \pm 354$ & $440 \pm 87$ \\ 
DMCheetahRun & $825 \pm 44$ & $\mathbf{856 \pm 25}$ & $822 \pm 13$ & $643 \pm 166$ & $3 \pm 1$ \\
Ant & $2273 \pm 1107$ & $\mathbf{3526 \pm 1162}$ & $2289 \pm 181$ & $3968 \pm 801$ & $2645 \pm 1462$ \\
HalfCheetah & $10758 \pm 397$ & $9764 \pm 583$ & $10358 \pm 709$ & $10100 \pm 963$ & $\mathbf{11451 \pm 406}$ \\
RoboAnt & $\mathbf{3041 \pm 137}$ & $2586 \pm 159$ & $1935 \pm 477$ & $2656 \pm 33$ & $974 \pm 136$ \\
RoboHalfCheetah & $836 \pm 188$ & $780 \pm 119$ & $824 \pm 54$ & $\mathbf{1713 \pm 582}$ & $681 \pm 134$ \\
Ant(B) & $\mathbf{3145 \pm 235}$ & $2555 \pm 280$ & $2101 \pm 412$ & $1962 \pm 1322$ & $808 \pm 29$ \\
HalfCheetah(B) & $\mathbf{2796 \pm 204}$ & $2209 \pm 78$ & $2135 \pm 195$ & $2676 \pm 57$ & $914 \pm 251$ \\
\bottomrule
\end{tabular}
\end{sc}
\end{small}
\vskip -0.1in
\label{table:summary}
\end{table}
\section{Conclusion}

We propose a framework which combines \gls{ES} with online hyper-parameter tuning of general off-policy learning algorithms. This framework extends the mathematical formulation of near on-policy based meta-gradients \citep{xu2018meta,zahavy2020self} and flexibly allows for the adaptation of both discrete and continuous variables. Empirically, this method  provides significant performance gains over static hyper-parameters in off-policy learning baselines. As part of the ongoing efforts in combining ES with off-policy learning, the current formulation greatly reduces the search space of the \gls{ES} subroutines, and makes the performance gains more consistent compared to prior work \citep{pourchot2018cem}.




\bibliographystyle{unsrt}
\bibliography{online_hyperparam_es}

\newpage
\section*{APPENDIX: Online Hyper-parameter Tuning in Off-policy Learning via Evolutionary Strategies}
\label{sec:appendix}

\subsection{Proof of Proposition 1}
\label{appendix:theory}
To show the equivalence, note first that the \gls{ES} gradient estimator is the REINFORCE gradient estimator \citep{williams1992} of the meta-objective. This gradient could be converted to its reparameterized gradient counterpart \citep{kingma2013auto} as follows
\begin{align}
    \mathbb{E}[\frac{1}{\sigma N} \sum_{j=1}^{N}\hat{L}_t^{(j)} (\eta_t^{(j)} - \mu_t)] &= \big[\nabla_\mu \mathbb{E}_{\eta_t \sim \mathbb{N}(\mu,\sigma^2)}[L(\psi_t + f(\psi_t,\mathcal{D},\eta_t),\eta_t)]\big]_{\mu=\mu_t} \nonumber \\
    &= \big[\nabla_\mu \mathbb{E}_{\epsilon\sim \mathcal{N}(0,1)}[L(\psi_t + f(\psi_t,\mathcal{D},\mu_t + \sigma\epsilon),\mu_t+\sigma\epsilon)]\big]_{\mu=\mu_t} \nonumber 
\end{align}
Then we expand the RHS in orders of $\sigma$. In particular, $L(\psi_t + f(\psi_t,\mathcal{D},\mu+\sigma\epsilon),\mu_t+\sigma\epsilon) = L(\psi_t+f(\psi_t,\mathcal{D},\mu_t),\mu_t) + \sigma\epsilon f_1 + \sigma^2\epsilon^2 f_2$, where $f_1,f_2$ are Taylor expansionns of the objective with respect to $\sigma\epsilon$. Due to the expectation, the first-order term vanishes due to $\mathbb{E}[\epsilon]=0$. And because we take the limit $\sigma\rightarrow 0$, the term with $f_2$ vanishes too.
When the meta-objective does not explicitly depend on the meta-parameter, i.e. $\nabla_\mu L(\psi,\mu) = 0$ (which is the case if the meta-objective is defined as the cumulative returns of the policy $\pi(\psi)$ as in \citep{xu2018meta,zahavy2020self} and our case), we finally have 
\begin{align*}
    \lim_{\sigma\rightarrow 0} \mathbb{E}[\frac{1}{\sigma N} \sum_{j=1}^{N}\hat{L}_t^{(j)} (\eta_t^{(j)} - \mu_t)]  = [\nabla_\psi L(\psi,\mu)]_{\psi=\psi_t+f(\psi_t,\mathcal{D},\mu)} [\nabla_\mu f(\psi_t,\mathcal{D},\mu)]_{\mu=\mu_t}. 
\end{align*}

\subsection{Experiment Details}
\label{appendix:exp}

\paragraph{Environment details.} We consider a set of similar tasks \emph{Walker}, \emph{Cheetah} and \emph{Ant} with different simulation backends: \texttt{Walker-v1}, \texttt{HalfCheetah-v1} and \texttt{Ant-v1} from OpenAI gym \citep{brockman2016}; \texttt{RoboschoolWalker-v1}, \texttt{RoboschoolHalfCheetah-v1} and \texttt{RoboschoolAnt-v1} from Roboschool \citep{klimov2017roboschool}; \texttt{WalkerRun}, \texttt{WalkerWalk}, \texttt{WalkerStand} and \texttt{CheetahRun} from DeepMind Control Suite \citep{tassa2018deepmind}; \texttt{Walker2dBullet-v0}, \texttt{HalfCheetahBullet-v0} and \texttt{AntBullet-v0} from Bullet Physics Engine \citep{coumans2010bullet}. Due to different simulation backends, these environments vary in several aspects, which allow us to validate the performance of algorithms in a wider range of scenarios. 

\paragraph{Normalization scores.} To calculate the normalization scores, we adopt the score statistics reported in Table \ref{table:statistics}. We summarize three statistics from $Z_t^{i,j}$ where $1\leq i\leq n$ indexes the algorithmic baseline, $1\leq j\leq m$ indexes the task and $t$ indexes the time tick during training. Three statistics are defined at each time tick $t$ and for each baseline $i$ as
\begin{align*}
    \text{mean}_t^i = \frac{1}{m} \sum_{j=1}^m Z_t^{i,j}, \text{median}_t^i = \text{median}(Z_t^{i,j}), \text{best ratio}_t^i = \frac{1}{m}\sum_{j=1}^m \mathbb{I}[i = \arg\max_{k} Z_t^{k,j}],
\end{align*}
where $\text{median}(Z_t^{i,j})$ refers to taking median across all task $1\leq j\leq m$ and $\mathbb{I}[\cdot]$ is the indicator function.

\paragraph{Implementation details.} The algorithmic baselines TD3, SAC and DDPG are all based on OpenAI Spinning Up \url{https://github.com/openai/spinningup} \citep{achiam2018openai}. We construct all algorithmic variants on top of the code base. To implement \gls{ES}-\gls{RL}, we borrow components from the open source code \url{https://github.com/apourchot/CEM-RL} of the original paper \citep{pourchot2018cem}.

\paragraph{Architecture.} All algorithmic baselines, including TD3, SAC and DDPG share the same network architecture following \citep{achiam2018openai}. The Q-function network $Q_\theta(x,a)$ and policy $\pi_\phi(x)$ are both $2$-layer neural network with $h=300$ hidden units per layer, before the output layer. All hidden layers have $\text{relu}$ activation. By default, for all algorithmic variants, both networks $Q_\theta(x,a),\pi_\phi(x)$ are updated with learning rate $\alpha=10^{-3}$. Other missing hyper-parameters take default values from the code base.

\subsubsection{Further implementation and hyper-parameter details}

Below we introduce the skeleton formulation and detailed hyper-parameter setup for each algorithmic variant. 

\paragraph{\gls{OHT}-\gls{ES} for discrete hyper-parameters.}
We focused on adapting the $n$-step hyper-parameter $\eta\equiv n$, which takes discrete values. The hyper-parameter is constrained to be $\eta\in \{1,2,...K\}$ for a total of $K$ values. We parameterize logits $L \in \mathbb{R}^K$ and update the distribution $P(\eta=i) \equiv \text{softmax}(L_i)$. 

As introduced in the main paper, we maintain $K$ agents, each corresponding to a hyper-parameter value $\eta$. At training iteration $t$, we sample $\eta\sim P$ and execute the corresponding agent. The trajectory is used for estimating the performance of the agent $\hat{L}_t^{(j)}$, then the logits are updated based on Eqn.(\ref{eq:discrete-esmeta}). To ensure exploration, when sampling the agent, we maintain a probability of $\epsilon=0.1$ to sample uniformly. The logits $L$ are initialized to be a zero-valued vector. We sample $N=6$ agents before carrying out updates on the logits. The update is with an Adam optimizer \citep{kingma2014adam} with learning rate $\beta=0.02$. To generate test performance, the algorithm samples from the distribution $P$ and evaluate its corresponding performance.

\paragraph{\gls{OHT}-\gls{ES} for continuous hyper-parameters.}

We focused on adapting the learning rate $\eta \equiv [\alpha_\pi,\alpha_q] = \alpha$. To ensure positivity of the learning rate, we take the parameterization $\alpha = 10^{\tilde{\alpha}}$ where $\tilde{\alpha} \in \mathbb{R}^2$ and update $\tilde{\alpha}$ with \gls{ES}.

At training iteration $t$, we sample $N=10$ perturbations of the current parameter means, with standard deviation $\sigma_{\tilde{\alpha}}$ initialized at $\in \{0.02,0.05, 0.2,0.5\}$, to be tuned based on each tasks. We find for most tasks, $\sigma_{\tilde{\alpha}}=0.5$ works the best. Note that though the standard deviation of $0.5$ is large in a typical \gls{ES} setting for \gls{RL} \citep{salimans2017evolution,choromanski2018structured,mania2018simple}, this induces relatively small changes in the space of $\alpha$. The mean of the Gaussian $\mu_{\tilde{\alpha}}$ is initialized at $[-3.0,-3.0]$ as we take the default learning rate for TD3 to be $\alpha=10^{-3}$ \citep{achiam2018openai}. We also maintain a main agent whose keeps training the policy parameter $\theta$ using the central learning rate $\alpha = 10^{\mu_{\tilde{\alpha}}}$ (this agent generates the test performance).

For the \gls{ES} update, we adopt \gls{CEM} \citep{de2005tutorial} for the update of $(\mu_{\tilde{\alpha}},\sigma_{\tilde{\sigma}})$. Note that unlike a gradient-based \gls{ES} update \citep{salimans2017evolution} here the standard deviation parameter $\sigma_{\tilde{\alpha}}$ is adjusted online. 

\paragraph{Meta-gradient for continuous hyper-parameters.}

Meta-gradients are designed for continuous hyper-parameters \citep{xu2018meta}. We still consider adapting the learning rates $\alpha$ as above. We elaborate how meta-gradients are implemented in details below.

In addition to the typical Q-function network $Q_\theta(x,a)$, we also train a meta Q-function critic $Q_{\text{meta}}(x,a) \approx Q^{\pi_\phi}(x,a)$ to approximate the Q-function of the current policy $\pi_\theta$. The meta-critic $Q_{\text{meta}}(x,a)$ has the same training objective as $Q_\theta(x,a)$ but they differ in predictions due to randomness in the initializations and updates. 

The meta objective is $L(\psi,\eta) \equiv \mathbb{E}[Q^\pi_{\phi}(x,a)] \approx \mathbb{E}[Q_{\text{meta}}(x,a)]$ where the expectation is taken such that $x\sim \mathcal{D},a=\pi_\phi(x)$. This meta objective estimates the off-policy gradient objective \citep{degris2012,silver2014}. Since the policy network is updated according to Eqn.(\ref{eq:ddpg}), we calculate the meta-gradient as $\Delta \alpha = \nabla_\alpha \mathbb{E}[Q_{\text{meta}}(x,\pi_{\phi^\prime}(x))]$, to be computed via back-propagations.

At training iteration $t$, the meta critic $Q_{\text{meta}}(x,a)$ is updated with the same rules as $Q_\theta(x,a)$ but with different batches of data. Then the learning rate $\alpha$ are updated via meta-gradients with the Adam optimizer \citep{kingma2014adam} and learning rate $\beta \in \{10^{-4},10^{-5}\}$, tuned for each task. Throughout the training, only one agent is maintained and trained, and this single agent generates the test performance.

\paragraph{\gls{ES}-\gls{RL}.} We implement the \gls{CEM}-\gls{RL} algorithm \citep{pourchot2018cem} but with our TD3 subroutines for fair evaluations. For critical hyper-parameters in the algorithm, we take their default values from the paper \citep{pourchot2018cem}.

Let $\theta$ be the agent parameter, the algorithm maintains a Gaussian distribution $(\mu_\theta,\sigma_\theta^2)$ for the agent parameter. At each iteration $t$, the algorithm samples $N=10$ parameters $\theta \sim \mathcal{N}(\mu_\theta,\sigma_\theta^2)$ from the distribution and updates $k=5$ agents using TD3 gradient updates. Then each agent is executed in the environment and their fitness (or cumulative returns) are collected. Finally, the distribution parameters $(\mu_\theta,\sigma_\theta^2)$ are updated via cross-entropy methods \citep{de2005tutorial}. The mean agent $\mu_\theta$ generates the test performance.

\begin{table}[t]
\centering
\caption{Summary of the performance scores used for calculating normalized scores across different simulated tasks. Normalized scores are calculated as $Z = \frac{R-L}{U-L}$ where $R$ is the test performance of a given baseline algorithm. Below, the low score $L$ is estimated by executing random policy in the environment. The high score $U$ is estimated as high performing returns in the selected tasks from related prior literature.}
\begin{small}
\begin{sc}
\begin{tabular}{C{1.in} *2{C{.8in}}}\toprule[1.5pt]
\bf Tasks & \bf High score $U$ & \bf Low score $L$ \ \\\midrule
DMWalkerRun & $1000$ & $0$  \\ 
DMWalkerWalk & $1000$ & $0$  \\ 
DMWalkerStand & $1000$ & $0$  \\ 
DMCheetahRun & $1000$ & $0$  \\ 
Ant & $6000$ & $-55$  \\ 
HalfCheetah & $10000$ & $-290$  \\ 
RoboAnt & $2500$ & $53$  \\ 
RoboHalfCheetah & $3000$ & $2$  \\
RoboWalker2d & $2500$ & $16$  \\
Ant(B) & $2500$ & $372$  \\
HalfCheetah(B) & $3000$ & $-1272$  \\
Walker2d(B) & $2500$ & $17$  \\
\bottomrule
\end{tabular}
\end{sc}
\end{small}
\vskip -0.1in
\label{table:statistics}
\end{table}

\end{document}